\documentclass[sigconf, screen]{acmart}
\AtBeginDocument{%
  }

\copyrightyear{2026}
\acmYear{2026}
\setcopyright{cc}
\setcctype{by}
\acmConference[MM '26]{Proceedings of the 34th ACM International Conference on Multimedia}{November 10--14, 2026}{Rio de Janeiro, Brazil}
\acmBooktitle{Proceedings of the 34th ACM International Conference on Multimedia (MM '26), November 10--14, 2026, Rio de Janeiro, Brazil}
\acmDOI{10.1145/3767308.3834748}
\acmISBN{979-8-4007-2213-4/2026/11}
\usepackage{balance}
\usepackage{multirow}
\usepackage{pifont}
\usepackage{textcomp}
\newcommand{\cmark}{\ding{51}}
\newcommand{\xmark}{\ding{55}}
\begin{document}

\title{OpenSAL360: Open-Source Crowdsourcing Platform for Omnidirectional Video Saliency Collection}

\author{Alexey Bryncev}
\orcid{0009-0003-9598-2376}
\authornote{Authors contributed equally to this research.}
\correspondingauthor
\affiliation{
  \institution{AI Center, Lomonosov Moscow State University}
  \department{MSU Institute for Artificial Intelligence}
  \city{Moscow}
  \country{Russia}
}
\email{alxbrc0@gmail.com}
\author{Andrey Moskalenko}
\orcid{0000-0003-4965-0867}
\authornotemark[1]
\affiliation{
  \institution{AI Center, Lomonosov Moscow State University}
  \department{MSU Institute for Artificial Intelligence, FusionBrain Lab}
  \city{Moscow}
  \country{Russia}
}
\email{and.v.moskalenko@gmail.com}
\author{Kira Shilovskaya}
\orcid{0009-0004-1123-1784}
\authornotemark[1]
\affiliation{
  \institution{Lomonosov Moscow State University}
  \city{Moscow}
  \country{Russia}
}
\email{kira.shilovskaya@graphics.cs.msu.ru}
\author{Ivan Kosmynin}
\orcid{0009-0000-1552-7583}
\affiliation{
  \institution{Lomonosov Moscow State University}
  \city{Moscow}
  \country{Russia}
}
\email{ivan.kosmynin@graphics.cs.msu.ru}
\author{Dmitriy Vatolin}
\orcid{0000-0002-8893-9340}
\affiliation{
  \institution{MSU Institute for Artificial Intelligence}
  \city{Moscow}
  \country{Russia}
}
\email{dmitriy@graphics.cs.msu.ru}
% \renewcommand{\shortauthors}{Bryncev A., Moskalenko A., Shilovskaya K. et al.}

\begin{abstract}
Omnidirectional video saliency prediction plays an important role in many immersive multimedia applications, including viewport-adaptive streaming and compression, foveated rendering, mesh simplification, perceptual quality assessment. Yet progress in this area remains constrained by the cost and complexity of collecting eye-tracking data with VR headsets, which makes large-scale dataset creation difficult to extend.
We present OpenSAL360, the first open-source platform for scalable, low-cost 360° video saliency collection. Unlike conventional VR-based protocols, it requires only a standard screen, mouse, and internet connection, enabling parallel saliency data collection from common crowdsourcing assessors without specialized hardware.
We validate our collection protocol against seven well-established VR eye-tracking datasets and conduct ablation studies on key interface, pre-, and post-processing parameters. To demonstrate the effectiveness and scalability of the proposed methodology, we collect and publicly release a saliency dataset covering 500 omnidirectional videos annotated by 2,000+ crowdsourcing assessors, making it, to the best of our knowledge, the largest dataset in this field. We make OpenSAL360 publicly available at \url{https://github.com/msu-video-group/OpenSAL360}.
\end{abstract}



%   \caption{Methodology overview.}


\maketitle

\begin{figure*}[t]
\centering
\includegraphics[width=0.962\textwidth]{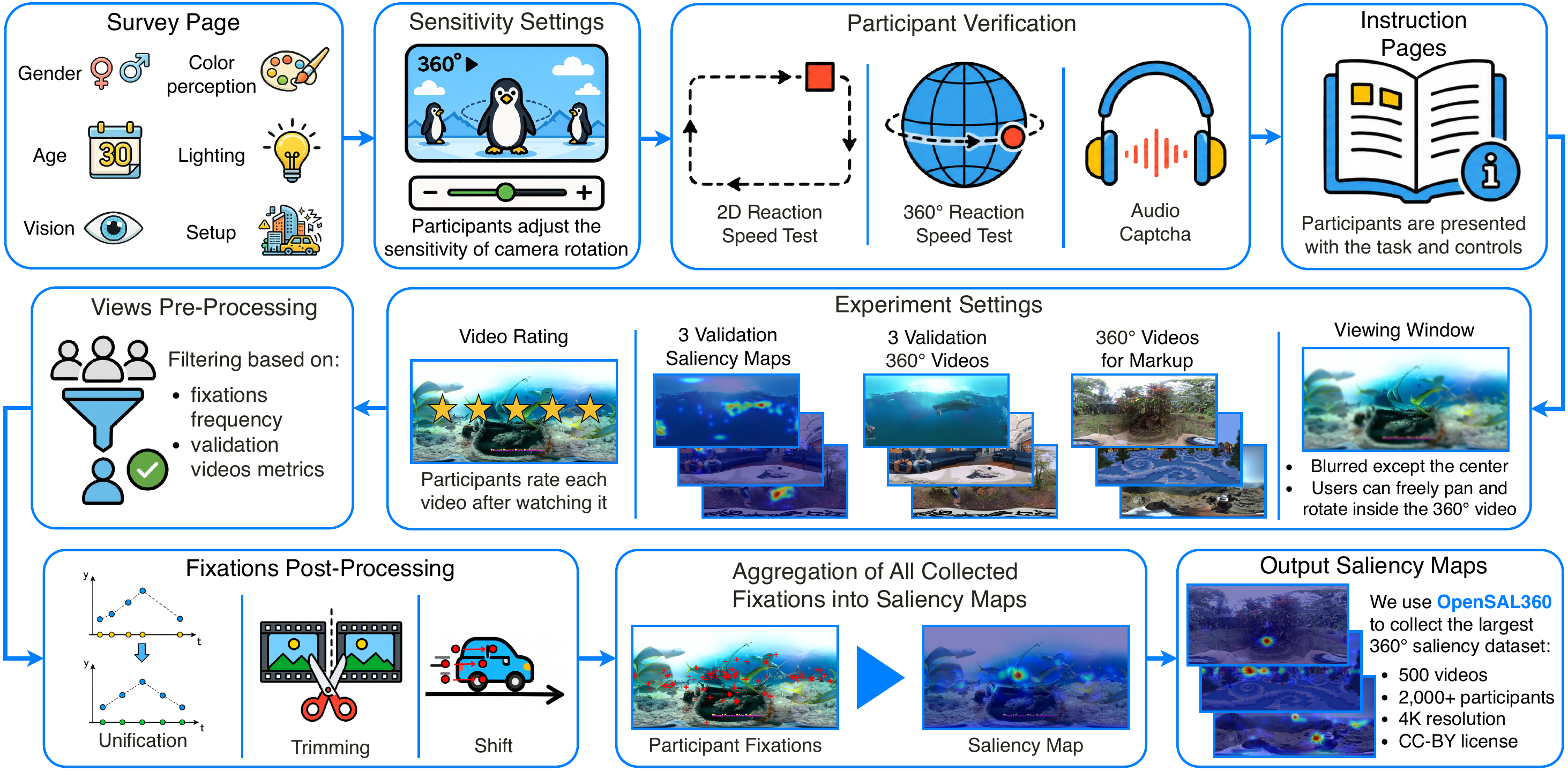}
\caption{OpenSAL360 pipeline for scalable omnidirectional video saliency collection using crowdsourcing.}
\Description{OpenSAL360 pipeline for scalable omnidirectional video saliency collection using crowdsourcing.}
\label{fig:pipeline}
\end{figure*}

\section{Introduction}

Visual saliency prediction aims to emulate the human visual system (HVS) by estimating how visual attention is distributed across a scene. In omnidirectional video, unlike fixed 2D viewing, observers actively navigate a spherical space, selecting viewports in real time.

Accurate saliency estimation plays an important role in a wide range of immersive multimedia applications including viewport-adaptive streaming~\cite{ozcinar2018visual,wang2022salientvr,pang2023vatp360,wang2024rosal360}, saliency-aware compression~\cite{baek2020sali360,ozcinar2021delivery,chiang2021saliency,gungordu2024saliency}, foveated rendering~\cite{patney2016towards,fan2024scene,foveated},  video quality assessment~\cite{li2018bridge,livqa,vqacnn,yang2024saliencyvqa,alexey2025bridging}, 3D mesh simplification~\cite{lee2005mesh,nousias2023deep,dos2023saliency}, cinematography and viewport guidance~\cite{su2016pano2vid,lai2017semantic,du2021saliency}, VR stitching~\cite{lin2017foveated}, and even privacy analysis of VR viewing behavior~\cite{nguyen2024penetration}. 

The close coupling between gaze fixation behavior, scene content, and human attention has made eye movements a natural and powerful signal for saliency modeling~\cite{yarbus,parkhurst2002modeling}. Progress in omnidirectional saliency prediction has been driven largely by datasets collected using head-mounted displays and eye-tracking systems~\cite{xu2018predicting,david2018dataset,xu2018gaze}. However, acquiring such datasets is inherently demanding: it requires specialized hardware, careful calibration, and controlled laboratory conditions. At the same time, the broader 2D saliency research has shown that attention-related supervision can be collected at scale with browser-based and mouse-contingent paradigms, as demonstrated for images and videos~\cite{salicon,bubbleview,focalvid}. Inspired by this direction, we present the OpenSAL360 platform, combining a 360° player, mouse-driven viewport control, and blurred-viewing interface that guides assessors to perceptually important regions.

Overall, our main contributions are as follows:

\begin{itemize}
    \item We introduce OpenSAL360, the first open-source crowdsourcing platform for omnidirectional video saliency collection that enables large-scale proxy fixations annotation without specialized VR headsets with eye-tracking.

    \item We provide a systematic validation of the data collection protocol by analyzing the impact of interface design, participant quality control, and pre-/post-processing choices. We show that the resulting annotations achieve strong agreement with well-established VR eye-tracking datasets.

    \item We demonstrate the scalability of OpenSAL360 by collecting and publicly releasing, to the best of our knowledge, the largest omnidirectional video saliency dataset to date. The dataset contains 500 videos and saliency annotations from 2,109 crowdsourcing assessors. The entire collection was completed in less than one week at a cost lower than that of a single consumer-grade VR eye-tracking headset.
\end{itemize}

\section{Related Work}
\subsection{Crowdsourced Video Saliency Collection}

Crowdsourcing has been widely used as a scalable alternative to laboratory eye tracking for saliency annotation in conventional 2D images and videos. Instead of recording gaze directly, mouse-based interfaces~\cite{lyudvichenko2019predicting,tavakoli2017saliency} ask participants to reveal blurred content, click on salient regions, or continuously follow visually important objects, as in SALICON~\cite{salicon}, BubbleView~\cite{bubbleview}, and FocalVid~\cite{focalvid}. Recent large-scale 2D video saliency prediction challenges~\cite{moskalenko2024aim,moskalenko2026ntire} further demonstrate that crowdsourced mouse annotations can support large-scale model training and reliable evaluation. Omnidirectional video introduces an additional challenge: human attention is naturally split between two components, namely the viewport selected on the sphere and the gaze location within the currently visible field of view. Therefore, directly transferring 2D saliency collection protocols to omnidirectional video is insufficient. This leaves a gap between scalable crowdsourced saliency collection for 2D video and high-fidelity but expensive VR eye-tracking studies.

\subsection{Omnidirectional Video Saliency Datasets}
The growing popularity of VR has increased interest in human attention, leading to the creation of several omnidirectional video saliency datasets summarized in Tab.~\ref{tab:all_datasets}. We validated OpenSAL360 on seven public VR eye-tracking datasets: two video-only datasets, PVS-HMEM~\cite{xu2018predicting} and Salient360!~\cite{david2018dataset}, and five audio-visual datasets, VR-EyeTracking~\cite{xu2018gaze}, 360AV-HM~\cite{icmew2020}, SVGC-AVA~\cite{yang2023svgc}, D-SAV360~\cite{bernal2023d}, and AVS-ODV~\cite{zhu2023audio}. Collecting such datasets requires VR headsets with integrated eye trackers, controlled laboratory conditions, and substantial manual effort, making large-scale collection difficult. The largest publicly available dataset contains only 208 videos~\cite{xu2018gaze}. In contrast, OpenSAL360 enables scalable, low-cost saliency collection using only a screen, mouse, and internet connection. With OpenSAL360, we collected saliency annotations for 500 CC-BY licensed YouTube omnidirectional 20s videos at low cost. Participants completed a session of 23 videos in under 18 minutes on average.

\begin{table*}[t!]
\centering
\fontsize{9pt}{10.255pt}\selectfont
\tabcolsep=2.538pt
\begin{tabular}{lccccccc}
\hline
\textbf{Dataset (Year)}
& \begin{tabular}[c]{@{}c@{}}\textbf{Video}\\ \textbf{Sources}\end{tabular}
& \begin{tabular}[c]{@{}c@{}}\textbf{Tracking}\\ \textbf{Sources}\end{tabular}
& \begin{tabular}[c]{@{}c@{}}\textbf{Unique}\\ \textbf{Videos}\end{tabular}
& \begin{tabular}[c]{@{}c@{}}\textbf{Duration}\\ \textbf{(s)}\end{tabular}
& \begin{tabular}[c]{@{}c@{}}\textbf{Participants}\\ \textbf{All/Per Video}\end{tabular}
& \textbf{Resolution}
& \textbf{Audio} \\ \hline
PVS-HMEM (2018)~\cite{xu2018predicting}            & YouTube/VRCun                                           & HTC Vive/aGlass2                                         & 76           & 10--80                                                 & 58 / 58                                                                & 3K--8K           & \xmark \\
Salient360! (2018)~\cite{david2018dataset}        & YouTube                                                 & HTC Vive/SMI                                             & 19           & 20                                                     & 57 / 57                                                                & 3840$\times$1920               & \xmark \\
VR-EyeTracking (2018)~\cite{xu2018gaze}      & YouTube                                                 & HTC Vive/aGlass                                          & 208          & 20--60                                                 & 45 / 31                                                                & 4K               & \cmark \\
VQA-ODV (2018)~\cite{li2018bridge}             & YouTube                                                 & HTC Vive/aGlass                                          & 60           & 10--23                                                 & 221 / 23                                                               & 4K--8K           & \cmark \\
SD360 (2018)~\cite{SD360}               & Sports-360                                              & HTC Vive/aGlass                                          & 104          & 20--60                                                 & 27 / 20                                                                & 4K               & \xmark \\
360\_EM (2019)~\cite{360_EM}             & YouTube                                                 & FOVE VR headset                                            & 15           & 38--60                                                 & 13 / 13                                                                & 1K--4K           & \cmark \\
360AV-HM (2020)~\cite{icmew2020}            & YouTube                                                 & Oculus Rift (HM as fixations)                                               & 21           & 25                                                     & 45 / 15                                                                & 3840$\times$1920 & \cmark \\
CEAP-360VR (2021)~\cite{xue2021ceap}          & YouTube                                                 & HTC Vive Pro Eye                                           & 8            & 60                                                     & 32 / 32                                                                & 3840$\times$1920 & \cmark \\
EHTask (2021)~\cite{ehtask}              & Public Datasets                                         & HTC Vive/7invensun                                       & 15           & 150                                                    & 30 / 30                                                                & 3840$\times$2160 & \xmark \\
VREED (2021)~\cite{vreed}               & YouTube                                                 & FOVE VR headset                                            & 12           & 60--180                                                & 43 / 34                                                                & 2048$\times$1080 & \cmark \\
QoE-360 (2022)~\cite{qoe360}             & JVET                                                    & HTC Vive Pro Eye                                           & 6            & 20                                                     & 31 / 31                                                                & 3840$\times$1920               & \xmark \\
D-SAV360 (2023)~\cite{bernal2023d}            & YouTube                                                 & HTC Vive Pro Eye                                           & 85           & 30                                                     & 87 / 55                                                                & 3840$\times$1920 & \cmark \\
AVS-ODV (2023)~\cite{zhu2023audio}             & Own videos                                              & HTC Vive Pro Eye                                           & 162          & 15                                                     & 60 / 20                                                                & \textbf{7680$\times$3840} & \cmark \\
SVGC-AVA (2024)~\cite{yang2023svgc}            & YouTube                                                 & HTC Vive Pro Eye/Droolon F1                              & 57           & 28                                                     & 63 / 63                                                                & 3840$\times$1920 & \cmark \\
Panonut360 (2024)~\cite{panonut360}          & YouTube                                                 & HTC Vive Pro Eye                                           & 15           & \textbf{140--352}                                      & 50 / 50                                                                & 3840$\times$1920 & \cmark \\
YT360-ET (2025)~\cite{cokelek2025spherical}   & YouTube                                                 & HTC Vive Pro Eye/Tobii                                   & 81           & 30                                                     & 102 / 15                                                               & 3840$\times$1920 & \cmark \\ \hline
\textbf{OpenSAL360 (2026)} & YouTube                                                 & Crowdsourcing                                              & \textbf{500} & 18.2                                                     & \textbf{2,109 / 84}                                                   & 3840$\times$1920 & \cmark \\ \hline
\end{tabular}
\caption{Comparison of existing omnidirectional video saliency datasets. Our dataset, collected using the OpenSAL360 platform, provides substantially larger video and observer coverage through scalable crowdsourcing.}
\label{tab:all_datasets}
\end{table*}

\subsection{Omnidirectional Video Saliency Models}
Early research on omnidirectional saliency relied mainly on handcrafted heuristics~\cite{bogdanova2008visual,bogdanova2010dynamic,fang2018novel} or adapted conventional 2D saliency models to spherical content by applying them to different projections and merging the predictions on the sphere~\cite{de2017look,lebreton2018gbvs360,chao2018salgan360,cheng2018cube}.

The emergence of omnidirectional saliency datasets enabled a shift toward data-driven models that better account for spherical geometry and temporal dynamics. ATSal~\cite{dahou2021atsal} combines global attention with local temporal modeling, while SD360~\cite{SD360} and Cube Padding~\cite{cheng2018cube} introduce geometry-aware processing to reduce projection distortions and boundary artifacts. SST-Sal~\cite{bernal2022sst} uses ConvLSTM and optical flow to capture temporal attention dynamics. Transformer-based approaches further improve long-range modeling: PAVER~\cite{yun2022paver} employs a ViT with deformable convolution to model relations among panoramic patch features, while SalViT360~\cite{cokelek2025spherical} adapts transformer attention to spherical geometry.

Audio-visual models improve saliency prediction by using sound cues, including off-screen audio. AVS360~\cite{chao2020towards} fuses audio and visual features, while~\cite{zhu2023unified} combines sound localization with semantic audio information. AViSal360~\cite{bernal2024avisal360} extends SST-Sal~\cite{bernal2022sst} with an audio branch, SVGC-AVA~\cite{yang2023svgc} models audio-visual relations on spherical graphs, and SalViT360-AV~\cite{cokelek2025spherical} integrates audio through transformer adapters. The growing scale of transformer-based models increases the need for larger, more diverse datasets and scalable data collection tools such as OpenSAL360.

\section{OpenSAL360 Open-Source Platform}
\subsection{Web User Interface}
We propose to collect omnidirectional video saliency through a web interface with a 360° video player, mouse-based viewport navigation, and continuous interaction logging (Fig.~\ref{fig:pipeline}). Videos are shown in full-screen mode, and, following mouse-based 2D saliency protocols~\cite{salicon,bubbleview,focalvid,lyudvichenko2019predicting,moskalenko2024aim,moskalenko2026ntire}, the interface uses blurred viewing: only a small region around the center remains sharp, while the rest is uniformly blurred. This encourages participants to keep salient content in focus, treating viewport movements as saliency cues.

We evaluated several interaction schemes: drag-and-drop and continuous mouse control with the focus region at the viewport center, as well as dynamic mouse-keyboard and mouse-edge navigation with the focus region at the cursor position. For each mode, we searched over the sharp region size $\sigma_{s}$ and field of view ($FOV$), then selected the blur level $\sigma_{b}$ on seven validation datasets using the average rank of 360° saliency metrics~\cite{david2024salient360} against eye-tracking ground truth. These experiments define the default preset: a drag-and-drop interface with $(\sigma_{s}, FOV, \sigma_{b})=(0.2, 40^\circ, 0.005)$, where $\sigma_{s}$ and $\sigma_{b}$ are measured relative to screen width.

Before annotation, each participant completes a demographic and viewing-condition survey, screen-size calibration using a ruler, credit, or business card~\cite{yung2015methods}, completes a viewing-instructions page, 2D and 360° reaction-time tests, and viewport sensitivity adjustment. In the main experiment, participants annotate 20 random videos and 3 honeypot videos with ground-truth eye-tracking saliency maps, and rate each video following~\cite{diem} to improve engagement.

\subsection{Architecture and Configurable Design}
OpenSAL360 is a configurable platform for large-scale omnidirectional saliency collection. Its admin panel supports video upload, regular and validation stimuli, experiment creation, and settings such as field of view, focus size, blur, audio, ratings, initial orientation, video count, and target views counter. A default preset derived from our ablations enables out-of-the-box usage.

The platform separates the participant web client, Django backend, and offline fixation processing. The HTML5/JavaScript client renders spherical video and records timestamped viewport orientations; the backend stores experiment and participant data, while a separate pipeline performs filtering, validation, temporal alignment, and saliency-map generation. All components are Dockerized.

Concurrent participants are isolated through server-side sessions and separate database records. Video assignment balances sampling to reach the target views counter. The modular design supports new interaction modes, participant checks, processing filters, and saliency-generation methods. The platform is not tied to any particular crowdsourcing provider. After completing the task, each participant receives a unique verification code, which can be used to identify the submission and process payment through external services such as Amazon Mechanical Turk~\cite{amt}, Toloka.ai, etc.

\begin{figure}[!tp]
  \centering
  \begin{minipage}{0.333\linewidth}
    \centering
    \includegraphics[width=\linewidth]{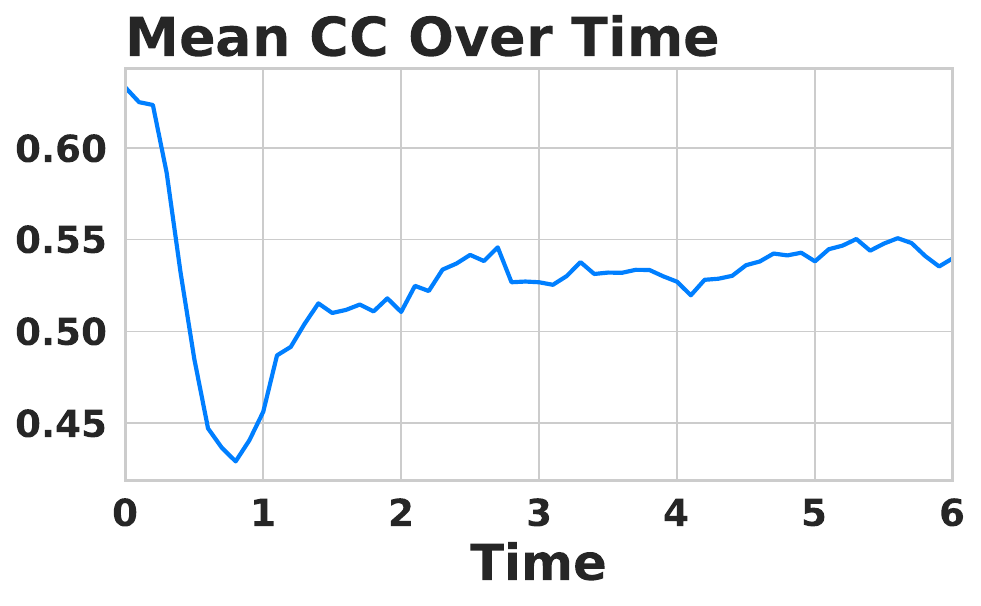}
  \end{minipage}\hfill
  \begin{minipage}{0.333\linewidth}
    \centering
    \includegraphics[width=\linewidth]{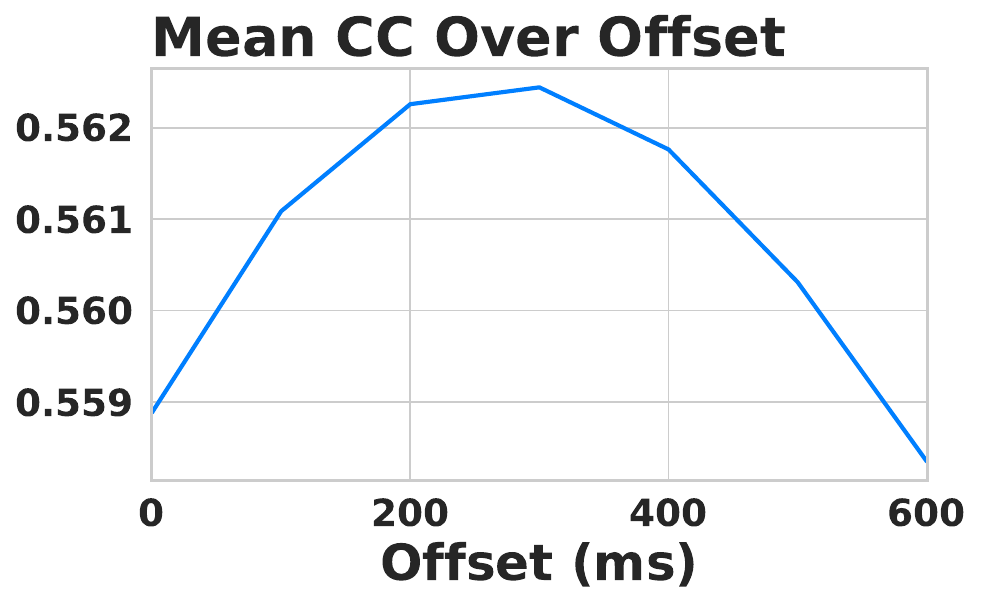}
  \end{minipage}\hfill
  \begin{minipage}{0.333\linewidth}
    \centering
    \includegraphics[width=\linewidth]{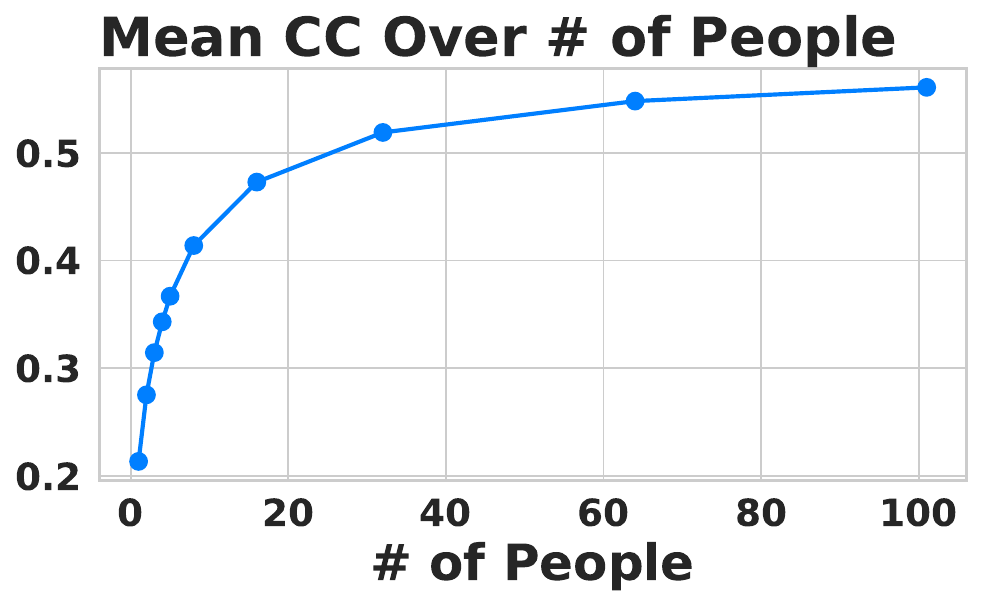}
  \end{minipage}
   \caption{Average CC across seven validation datasets for selecting video trimming time (left), temporal offset (middle), and number of crowdsourced observers per video (right).}
   \Description{Average CC across seven validation datasets for selecting video trimming time (left), temporal offset (middle), and number of crowdsourced observers per video (right).}
  \label{fig:postproc}
\end{figure}

\subsection{Data Collection Pre-Processing}
Before annotation, OpenSAL360 ensures reliable viewing conditions and participant readiness. To avoid network interruptions, all videos are fully preloaded, displayed in full-screen mode, and resized to the largest possible area while preserving aspect ratio. Since our videos contain audio, we verify that audio is enabled three times during the experiment using audio captchas. Participants with screens below 1280$\times$720 are excluded to avoid artifacts from heavy downscaling.

Viewport-control accuracy is verified with two reaction-time tests. In the 2D test, repeated three times, a white square moves along the screen boundary and participants must keep the cursor inside the square. They pass if the cursor remains inside the square for at least 25\% of the time in at least two attempts. After choosing their preferred 360° sensitivity on an unblurred demo video, participants complete a two-attempt omnidirectional test, where a circular target moves across a large portion of the sphere. They pass if the viewport center remains inside the target for at least half of each attempt. These checks ensure that contributors understand the interface and can reliably control the viewport before annotation.

\subsection{Data Collection Post-Processing}
In addition to the 20 target videos, each participant annotates 3 honeypot videos with reference eye-tracking saliency maps. We generate saliency maps from these participant views and compare them with the reference using the CC metric~\cite{david2024salient360, bylinskii2018different}. Honeypots were selected to have average CC values near the 50th percentile on the validation datasets, avoiding clips that are too easy or too difficult. For each honeypot, the acceptance threshold was set to the 5th percentile of its validation CC distribution. Participants falling below threshold on any honeypot had all data discarded; this filtered out 16\% of participants, who were still compensated. The selected thresholds and ablations are provided in the OpenSAL360 Wiki.

We further discard recordings with event frequencies below 3Hz and resample the remaining trajectories to 100Hz. Before saliency-map generation, we shift cursor traces backward by 300ms to compensate for the delay between eye movements and mouse-controlled viewport motion and trim the first 1.5s of each trajectory. Initial correspondence with VR eye tracking is artificially high because VR eye-tracking viewers also start at the video center. 

We also analyzed metric behavior as the number of observers increased; saturation occurs at approximately 100 observers before filtering, which we set as the recommended value. Plots are provided in Fig.~\ref{fig:postproc}. All processed viewport-center trajectories are treated as proxy fixation annotations and converted into final saliency maps using Gaussian smoothing with an angular radius consistent with standard VR eye-tracking protocols~\cite{david2018dataset,bernal2023d,david2024salient360}.

Tab.~\ref{tab:comparison_360} compares the main OpenSAL360 ablation steps with state-of-the-art automatic omnidirectional video saliency prediction models. Our saliency data are substantially closer to reference VR eye-tracking data than predictions from any automatic model.

\begin{table}[!tp]
\fontsize{9pt}{10.846pt}\selectfont
\tabcolsep=8.15pt
\centering
\begin{tabular}{lccc|ccc}
\hline
\textbf{Method}
& 
& 
& 
& $\mathbf{CC}_{360}$
& $\mathbf{SIM}_{360}$
& $\mathbf{NSS}_{360}$ \\ \hline
\multicolumn{4}{l|}{PAVER~\cite{yun2022paver}}                                               & 0.2238                          & 0.2267                          & 1.0675                          \\ 
\multicolumn{4}{l|}{ATSal~\cite{dahou2021atsal}}                                             & 0.2337                          & 0.2564                          & 1.5945                          \\
\multicolumn{4}{l|}{SST-Sal~\cite{bernal2022sst}}                                            & 0.3213                          & 0.3157                          & 2.0526                          \\
\multicolumn{4}{l|}{ViNet~\cite{vinet} (ERP)}                                                & 0.3848                          & 0.3373                          & 2.0240                          \\
\multicolumn{4}{l|}{SVGC-AVA~\cite{yang2023svgc}}                                            & 0.4288                          & 0.3717                          & 2.2341                          \\
\multicolumn{4}{l|}{SalFoM~\cite{salfom} (ERP)}                                              & 0.4309                          & 0.3624                          & 2.3717                          \\
\multicolumn{4}{l|}{SalViT360~\cite{cokelek2025spherical}}                                   & 0.4597                          & 0.3650                          & 2.3973                          \\ \hline
\multicolumn{1}{l|}{\multirow{5}{*}{Ours}} & V                     & P                     & O                     &                                 &                                 &                                 \\ \cline{2-7} 
\multicolumn{1}{l|}{}                      & \xmark & \xmark & \xmark & 0.5089                          & 0.3884                          & 4.5895                          \\
\multicolumn{1}{l|}{}                      & \cmark & \xmark & \xmark & 0.5184                          & 0.3942                          & \textbf{4.7841}                 \\
\multicolumn{1}{l|}{}                      & \cmark & \cmark & \xmark & \underline{0.5220} & \underline{0.3992} & 4.4409                          \\
\multicolumn{1}{l|}{}                      & \cmark & \cmark & \cmark & \textbf{0.5585}                 & \textbf{0.4306}                 & \underline{4.7524} \\ \hline
\end{tabular}
\caption{OpenSAL360 vs. automatic methods across all validation datasets. Best and second-best results are shown in \textbf{bold} and \underline{underlined}; bottom rows ablate viewing parameters (V), post-processing (P), and 100 observers before filtering (O).}
\label{tab:comparison_360}
\end{table}

\section{Ethical Considerations and Privacy}
Data collection was conducted via \url{www.Toloka.ai}. We collected viewport movements as saliency traces from paid crowdsourcing workers and compensated them above the average local wage for the estimated task duration; we encourage future OpenSAL360 users to follow the same fair-pay policy. OpenSAL360 does not collect sensitive personal information: recorded trajectories and survey answers contain no information that can identify an individual. 

For the released dataset, we provide a removal mechanism for affected parties. Copyright holders or individuals appearing in the videos may contact us through the OpenSAL360 webpage; justified requests will be reviewed, and the corresponding videos removed.

\section{Conclusion}
In this paper, we introduced OpenSAL360, the first open-source crowdsourcing platform for omnidirectional video saliency collection. By carefully designing and validating the data collection pipeline against VR eye-tracking datasets, we showed that large-scale crowdsourcing can serve as a practical and effective proxy for conventional VR eye tracking in omnidirectional saliency annotation. Using OpenSAL360, we collected, to the best of our knowledge, the largest publicly available omnidirectional video saliency dataset to date, comprising 500 videos annotated by 2,109 observers, with over 84 views per video on average after filtering.

\begin{acks}
The work of Alexey Bryncev and Andrey Moskalenko was supported by the The Ministry of Economic Development of the Russian Federation in accordance with the subsidy agreement (agreement identifier 000000C313925P4H0002; grant No 139-15-2025-012). The research was carried out using the MSU-270 supercomputer of Lomonosov Moscow State University.
\end{acks}

\bibliographystyle{ACM-Reference-Format}
\balance
\bibliography{references}

% \appendix

\end{document}